\pdfoutput=1

\documentclass[11pt]{article}

\usepackage{acl}

\usepackage{times}
\usepackage{latexsym}

\usepackage[T1]{fontenc}

\usepackage[utf8]{inputenc}

\usepackage{microtype}

\usepackage{inconsolata}

\usepackage{graphicx}

\usepackage{hyperref}
\usepackage{tabularx}

\title{Negation-Induced Forgetting in LLMs}

\author{
 \textbf{Francesca Capuano},
 \textbf{Ellen Boschert},
 \textbf{Barbara Kaup}
\\
\\
 Schleichstra{\ss}e 4, 
	72076 T{\"u}bingen, Germany,\\
    Department of Psychology, 
    University of T{\"u}bingen\\
\\
 \small{
   \textbf{Correspondence:} \href{mailto:francesca.capuano@uni-tuebingen.de}{francesca.capuano@uni-tuebingen.de}
 }
}

\begin{document}

\maketitle

\begin{abstract}
The study explores whether Large Language Models (LLMs) exhibit negation-induced forgetting (NIF), a cognitive phenomenon observed in humans where negating incorrect attributes of an object or event leads to diminished recall of this object or event compared to affirming correct attributes \citep{mayo2014if, zang2023negation}. We adapted \citet{zang2023negation} experimental framework to test this effect in ChatGPT-3.5, GPT-4o mini and Llama3-70b-instruct.
Our results show that ChatGPT-3.5 exhibits NIF, with negated information being less likely to be recalled than affirmed information. GPT-4o-mini showed a marginally significant NIF effect, while LLaMA-3-70B did not exhibit NIF. 
The findings provide initial evidence of negation-induced forgetting in some LLMs, suggesting that similar cognitive biases may emerge in these models. This work is a preliminary step in understanding how memory-related phenomena manifest in LLMs.
\end{abstract}

\section{Introduction}

Recent advances in Natural Language Processing (NLP) and the widespread application of Large Language Models (LLMs), such as ChatGPT, have sparked increasing interest in understanding how these models process and retain information. A common research approach is to compare the linguistic and cognitive abilities of LLMs with those of human participants (\citealp[e.g.][]{lampinen2024language, binz2023using, cai2023does, hu2022fine}). Such comparative studies not only deepen our understanding of LLMs but also inform broader debates on human language processing, including memory and learning mechanisms \citep{chang2022word, warstadt2022artificial}.

One cognitive bias relevant to humans is negation-induced forgetting (NIF), in which negating incorrect information leads to greater memory impairment than affirming correct information. \citet{mayo2014if} first demonstrated this effect in human studies, showing that participants were more likely to forget details after negating incorrect attributes compared to affirming true ones. For example, when participants were asked whether someone had drunk red wine after witnessing them drink white wine, those who correctly responded "No" were more likely to forget that wine was consumed at all, compared to those who correctly responded "Yes" after being asked whether someone had drunk white wine. This suggests that negation disrupts memory consolidation, making it more difficult to retrieve information later. 
\citet{zang2023negation} extended this research by reducing associative interference and confirming that NIF persists even under controlled conditions. Their study involved a verification task, where participants answered yes/no questions about a story, followed by a free recall task. The key finding was that information from negated statements was less likely to be recalled than information from affirmed statements, supporting the NIF effect.

Whether LLMs replicate human cognitive biases—such as NIF—remains an open question. Prior studies suggest that LLMs exhibit well-known cognitive heuristics, such as the anchoring effect and framing effect, even though these biases are typically associated with human cognition (\citealp[e.g.][]{nguyen2024human, suri2024large, schubert2024context}). If LLMs also exhibit NIF, this could indicate that memory biases can potentially emerge from their training routine, without the need to explicitly integrate human-like reasoning mechanisms and controlled processes such as active inhibition.

By adapting Zang et al.’s methodology and comparing the performance of different conversational AI models to human results, this study contributes to ongoing debates about whether LLMs replicate human cognitive biases and how memory-related phenomena are encoded in large-scale language models. Understanding whether NIF occurs in LLMs can inform both theoretical models of linguistic cognition and practical considerations for AI deployment.

If LLMs display negation-induced forgetting effects, this may be attributed to inherent characteristics of LLMs, such as the attention mechanisms in the transformer architecture, which might distribute focus differently for information appearing in the context of negation, reducing its prominence in subsequent processing. Additionally, the statistical patterns learned during training could inherently weight negated information differently, leading to apparent "forgetting".

\section{Pilot}

We collected preliminary data in a pilot study conducted in the context of a psychology course at the University of Tuebingen.

\subsection{Participants}
The original sample comprised 40 ChatGPT-3.5 conversations, accessed via  \href{https://openai.com/}{OpenAI}. Each chat was treated as a participant. 

\subsection{Materials}

The experimental setup and materials were adapted from \citet{zang2023negation}. We used their short story about two university students, Montse and Jordi. The story consisted of 62 sentences, of which 44 were experimental sentences, 8 were fillers and the remaining 10 were baseline sentences. 
Experimental sentences were verified, filler sentences were sometimes verified but were not included in the analysis, baseline sentences were not verified but served as control condition (see below for more details on the procedure).
The story was presented in either one of two forms, counterbalancing the polarity of the verified sentences.

\begin{table}[]
    \centering
\begin{tabular*}{\linewidth}{@{\extracolsep{\fill}} l }
    \hline
         \textbf{Study Phase (excerpt)}  \\
         \textit{[...] Montse is a psychology student.} \\
         \textit{Montse has a cheerful character.[...]}\\
         \textit{Jordi is an informatics student.} \\
         \textit{Jordi has a confident character.[...]}\\
         \textit{After lunch, Montse and Jordi go to a workshop.}\\

         \textbf{Experimental Affirmed Sentence}  \\
         \textit{The person who is a psychology student is Montse.} \\
         Correct answer: \textit{Yes.} \\

         \textbf{Experimental Negated Sentence}  \\
         \textit{The person who is an informatics student is Montse.} \\
         Correct answer: \textit{No.} \\

         \textbf{Baseline Sentence}\\
         \textit{Montse has a cheerful character.}\\
        
        \textbf{Filler Sentence}\\
        \textit{After lunch, Montse and Jordi go to a workshop.}\\ 
        Correct answer: \textit{Yes.} \\
\hline
\end{tabular*}
     \caption{\textbf{Example items -} This is a demonstrative example: Experimental sentences are presented during the verification phase and can have affirmative or negative answers. Some filler sentences also appear in the verification phase, but they are not considered in the analysis. Baseline sentences are not verified and serve as the baseline condition for comparison with verified sentences.}
    \label{tab:my_label}
\end{table}

\subsection{Procedure}

The chats were treated as human participants. They were introduced with the prompt: “\textit{In the following, we would like to conduct a psychological experiment with you. Please read and follow the instructions below}”. Before the experiment, they were informed that they would read a story about two university students and subsequently answer questions about it.
The experiment consisted of four phases:

\begin{itemize}
    \item \textbf{Study Phase}: the chat was presented with the short story about two university students presented from a third-person perspective.    
    \item \textbf{Verification Task}: the chat was asked to decide whether each of the 22 presented statements accurately reflected the events in the story. The chat was asked to respond with either “Yes” or “No” to each statement. Only experimental and filler sentences were verified.
    \item \textbf{Filler Task} (Optional): If a filler task was included, the chat completed a Python coding task. In the no-filler condition, the free recall task followed immediately after the verification task. 
    \item \textbf{Free Recall Task} the chat was asked to recall as much as possible from the story.
\end{itemize}

Chat runs were randomly assigned either to the filler ($N=18$) or to the no-filler ($N=19$) condition. 

\subsection{Results}

\subsubsection{Verification Task}

We first assessed the accuracy of responses in the verification task for both the affirmative and negative conditions. Responses were hand-coded by four coders, each coding 10 chats. Randomly assigned parts of the coding were cross-checked for consistency.

A binomial generalized linear mixed-effects model (GLMM) was fitted to predict task accuracy with polarity, filler task, and their interaction as fixed effects. The model included subject and sentence number as random intercepts. A Type III Wald chi-square test was conducted to assess the significance of the fixed effects.
There was a significant main effect of polarity, ${\chi}^2(1)=5.85$, $p=.016$, indicating that polarity influenced task accuracy. However, the main effect of filler task was not significant, ${\chi}^2(1)=0.48$, $p=.49$, nor was the interaction between polarity and filler task, ${\chi}^2(1)=0.01$, $p=.91$.
Interestingly, accuracy was higher for negative ($M=83\%$, $SD=7\%$) compared to affirmative ($M=75\%$, $SD=13\%$) responses, which contrasts with the findings of \citet{zang2023negation}, where accuracy was similar across both response types (affirmative: $M = 86.9\%$, $SD = 8.7\%$; negative: $M = 85.3\%$, $SD = 9.6\%$; $t(34)=0.91$, $p=.37$).

\subsubsection{Free Recall}

For the free recall task, each chat's response was evaluated by comparing it to the original story. Each sentence from the story was split into components (character, verb, object), with a point assigned for each correctly recalled component (range: 0-3 points per sentence). These points were then used to calculate the proportion of memory failure for each participant and condition. This coding procedure followed the one used in \citet{zang2023negation}.
Only correct verification responses were included in the analysis. Following this criterion, 8.6\% of the observations were excluded.

A linear mixed-effects model was fitted to the data. The model included filler task, polarity, and their interaction as fixed effects, with subject and sentence number as random intercepts. The ANOVA table shows a significant main effect of polarity on memory failure, $F(1,647)=4.64$, $p=.03$, but no significant effect of filler task, $F(1,39)=1.49$, $p=.23$, and no significant interaction between filler task and polarity, $F(1,641)=0.65$, $p=.42$. 

Another model testing the effect of verification (baseline vs. experimental) on memory failure reveals no significant main effect of filler task, $F(1,40)=0.81$, $p=.37$, or sentence condition, $F(1,52)=1.61$, $p=.21$. The interaction between filler task and sentence condition was also non-significant, $F(1,1889)=0.39$, $p=.53$.

\begin{figure}
    \centering
    \includegraphics[width=\linewidth]{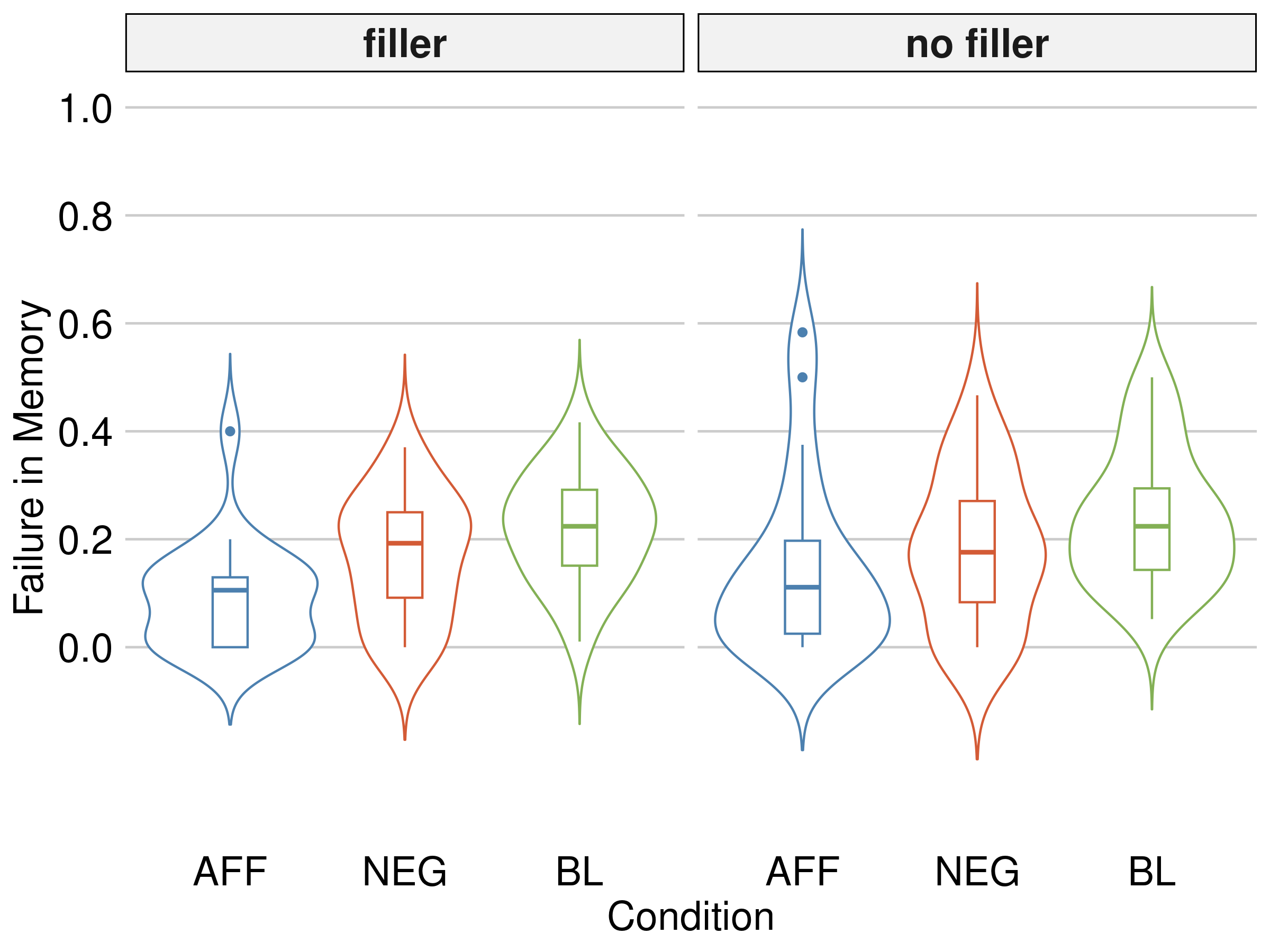}
    \caption{\textbf{Pilot -} Negation-induced forgetting effect in the filler (left) and in the no-filler (right) conditions.}
    \label{fig:pilot}
\end{figure}

Overall, the results support the hypothesis that ChatGPT-3.5 exhibits negation-induced forgetting effects. This effect was not influenced by the presence or absence of a filler task between the verification and free recall tasks (see Figure \ref{fig:pilot}).

\begin{figure}
    \centering
    \includegraphics[width=\linewidth]{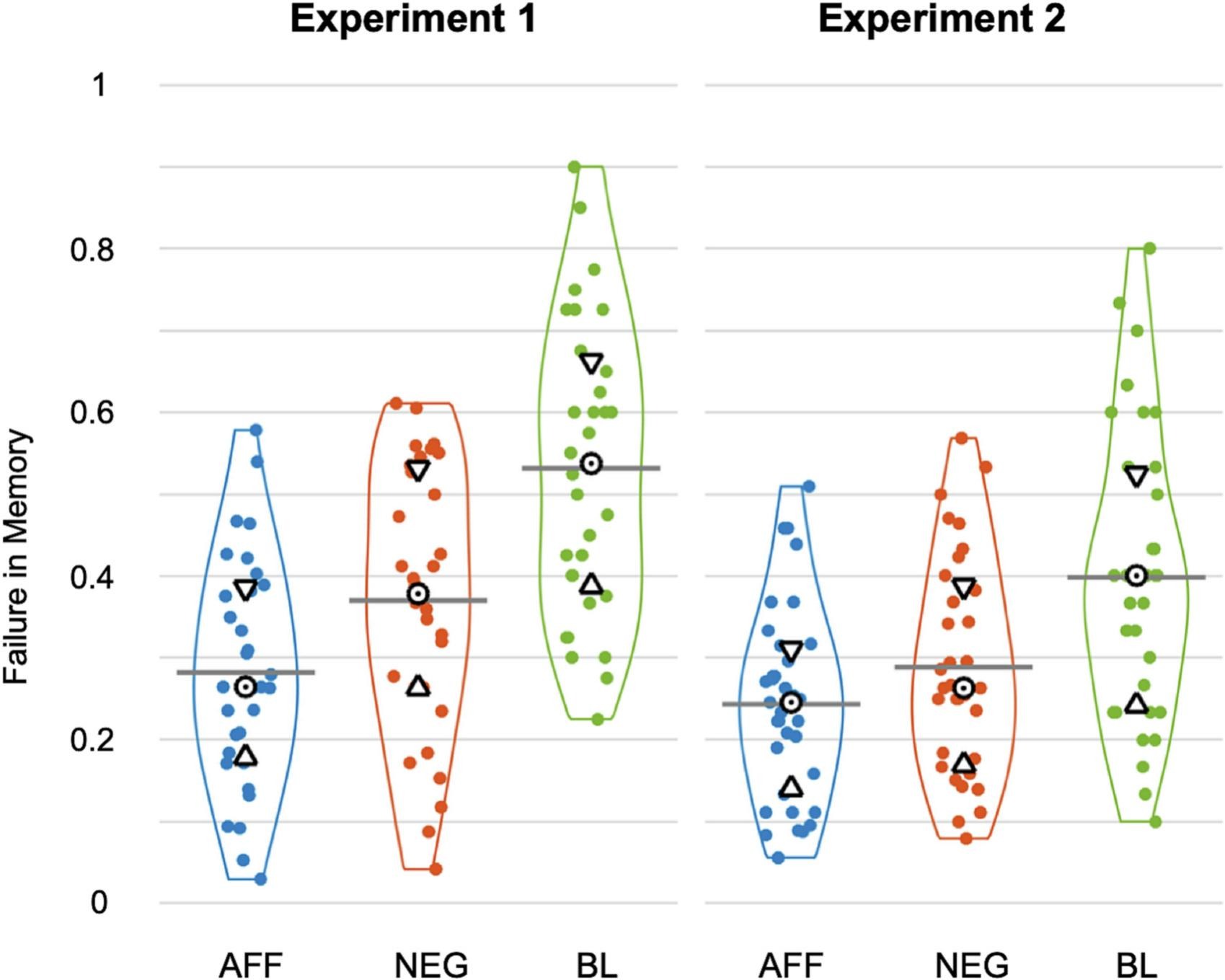}
    \caption{Negation-induced forgetting effect in \citet{zang2023negation}'s Experiments 1 and 2.}
    \label{fig:zang}
\end{figure}

\section{Main Experiments}

The pilot study was used to run a power analysis. At least 80 test runs were necessary to reliably detect a main effect of polarity (80\% power) of the magnitude of the one found in the pilot study. We decided to run 100 tests per experiment.  
These larger experiments involved GPT-4o-mini and LlaMA-3-70B and were pre-registered on \href{https://doi.org/10.17605/OSF.IO/QMFDW}{OSF}.

\subsection{Materials and Procedure}

Materials and procedure were the same as in the pilot, except for the following changes. Two versions of the story were created, differing only in the attribution of a key fact to one of the protagonists within a semantic pair. Each version was further presented in two counterbalanced forms, reversing the polarity of the verified sentences. 
The distraction task was always present and consisted in writing html code to implement the homepage of an online personal finance tracker. 

\subsection{GPT-4o-mini}

\subsubsection{Verification Task}

A binomial linear mixed model (GLMM) was used to examine the effect of Polarity on answer correctness in the verification task. 
The final model included the fixed effect of Polarity and random intercepts for both Subject and Item. 
The main effect of Polarity on the likelihood of correctness was not statistically significant ($\beta=0.872$, $p=.11$). In general, accuracy was much higher than in the pilot study (aff: $M=93\%$, $SD=7\%$; neg: $M=95\%$, $SD=5\%$).

\subsubsection{Free Recall}

\begin{figure}
    \centering
    \includegraphics[width=\linewidth]{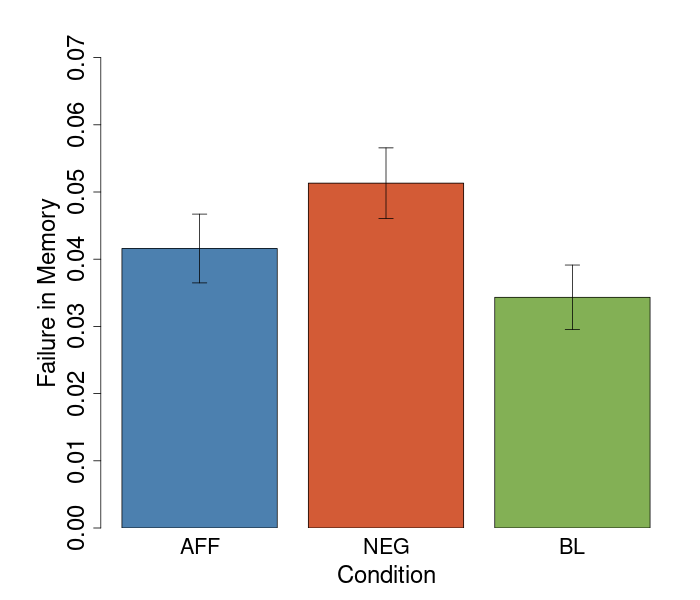}
    \caption{\textbf{GPT-4o-mini - } Mean failure in memory per condition $\pm$ standard error. The negation induced forgetting effect is marginally significant.
}
    \label{fig:gpt4}
\end{figure}

The chat was interrogated through the OpenAI API.
In general, the failures in memory are much less than those of ChatGPT-3.5 (see Figure \ref{fig:gpt4}). 
A linear mixed-effects model was fitted to examine the effect of polarity on failure rate, with Polarity as a fixed effect and Subject and Item as random effects, allowing for by-subject random slopes for Polarity. The effect of Negative Polarity was marginally significant ($\beta=0.011$, $p=.058$), suggesting a trend toward higher failure rates for negative polarity sentences.

A second linear mixed-effects model was fitted to compare the Experimental Condition with a baseline, including Subject as a random intercept. The effect of the Experimental Condition was not significant ($\beta=0.012$, $p=.092$), indicating no strong evidence of a difference in failure rates between baseline and experimental conditions.

\subsection{LlaMA-3-70B}

\subsubsection{Verification Task}

A binomial linear mixed model (GLMM) was used to examine the effect of Polarity on answer correctness, with random intercepts for Subject and Item. 
There was no significant effect of Polarity ($\beta=0.07$, $p=.87$). The mean accuracies are comparable to those of GPT-4o-mini (aff: $M=94\%$, $SD=7\%$; neg: $M=95\%$, $SD=5\%$).

\subsubsection{Free Recall}

\begin{figure}
    \centering
    \includegraphics[width=\linewidth]{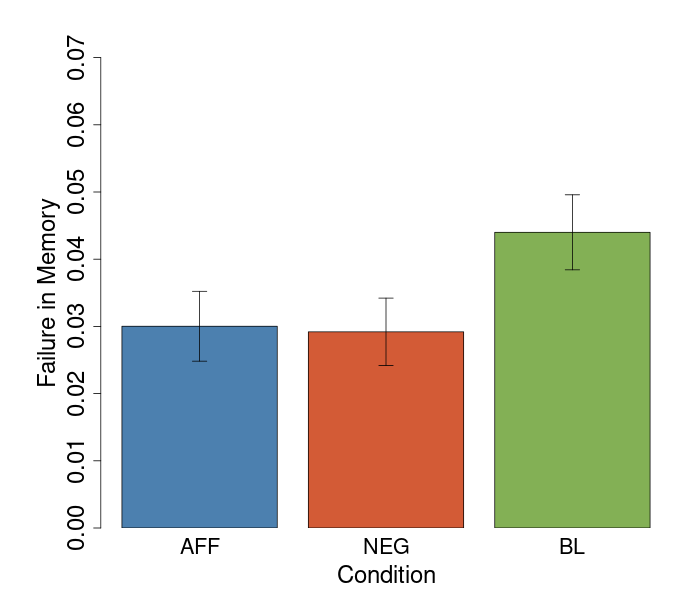}
    \caption{\textbf{LlaMA-3-70B -} Mean failure in memory per condition $\pm$ standard error. There is no negation-induced forgetting effect.
}
    \label{fig:llama}
\end{figure}

LlaMA-3 was accessed through the Replicate API.
The results can be seen in Figure \ref{fig:llama}. 
A linear mixed-effects model was fitted to examine the effect of Polarity on Free Recall, with Subject and Item as random intercepts. 
The analysis reveals no significant effect of Polarity ($\beta=0.001$, $p=.92$), indicating that polarity did not influence free recall performance.

The second linear mixed-effects model tested the effect of Experimental Condition on Free Recall and including Subject as a random intercept. The effect of Experimental Condition was significant ($\beta=0.044$, $p<.05$), suggesting that recall performance differed between conditions. Baseline sentences were rememebered worse than verified sentences.

\section{Data Availability}

Data and analysis scripts are available on \href{https://osf.io/dgjvk/?view_only=2dfab11454b049d2871ca61268412afb}{OSF}.

\section{Discussion and Conclusions}

Our study investigated whether Large Language Models (LLMs) exhibit negation-induced forgetting (NIF), a cognitive bias observed in human memory. By adapting the experimental design of \citet{zang2023negation}, we tested this effect in ChatGPT-3.5, GPT-4o-mini, and LLaMA-3-70B. Our findings suggest that, like humans, LLMs can exhibit NIF, with negated information being less likely to be recalled than affirmed information.

In our pilot study, ChatGPT-3.5 showed a significant main effect of polarity on memory failure, supporting the presence of NIF in LLMs. This effect persisted regardless of the inclusion of a filler task between the verification and recall phases, which was designed to mimic the distraction task used in human studies. The main experiments with GPT-4o-mini and LLaMA-3-70B only partially validated these findings: GPT-4o-mini exhibited a marginally significant NIF effect, while LLaMA-3-70B showed no such effect. However, the latter two models achieved extremely high accuracy with very low failure rates, suggesting possible ceiling effects.

Our results align with prior research showing that LLMs exhibit various cognitive biases. These biases might arise from the statistical learning mechanisms inherent in transformer-based architectures, which prioritize certain linguistic patterns over others. In the case of NIF, LLMs may allocate lower attention weights to information in the context of negation during encoding, leading to reduced recall accuracy. Further research is needed to examine how LLMs internally represent negation and whether these representations parallel human cognitive models.

The presence of NIF in LLMs has both theoretical and practical implications. Theoretically, our study contributes to ongoing discussions on whether LLMs replicate human memory biases and cognitive mechanisms. Our findings suggest that memory biases can emerge in LLMs, potentially suggesting the role of attention-based processing in shaping recall patterns. Practically, these findings highlight possible risks in AI applications: if LLMs exhibit NIF, they may selectively forget negated information, potentially compromising their reliability in tasks that require precise recall of factual content.

This study serves as an initial step in exploring NIF in LLMs. Future research should systematically examine how NIF varies across model sizes, architectures and training data distributions. Adapting the paradigm to simpler autoregressive LLMs could help determine whether even basic next-word prediction can account for NIF.
Additionally, investigating whether instruction tuning amplifies or mitigates this effect could offer insights into how training strategies shape memory biases in LLMs. 

In conclusion, our study provides evidence that LLMs exhibit negation-induced forgetting, mirroring human cognitive biases. These findings advance our understanding of how LLMs process negation and recall information, offering valuable insights for both cognitive science research and AI model development.

\section{Limitations}

One obvious limitation of our study is that we tested proprietary models (ChatGPT-3.5 and GPT-4o-mini), meaning their underlying architectures and training data are not fully accessible. The only open-source model investigated, Llama3, does not show a NIF effect. This restricts the generalizability of findings across different language models, and largely limits the possibility to later on directly assess the precise mechanisms that may be contributing to the observed effects, such as attention weights or training data biases. However, it seems unlikely that NIF was a desired feature directly engineered into the model. 
Future research could benefit from more transparent and reproducible models, which would allow for deeper insights into the underlying processes of negation-induced forgetting and other cognitive phenomena.

Secondly, the models tested in this study represent only a small subset of available LLMs, and their performance may not generalize to other models, especially those trained on different datasets or with different architectures. Further research involving a broader range of models will be necessary to draw more robust conclusions.

\bibliography{main}

\begin{thebibliography}{11}
\providecommand{\natexlab}[1]{#1}

\bibitem[{Binz and Schulz(2023)}]{binz2023using}
Marcel Binz and Eric Schulz. 2023.
\newblock Using cognitive psychology to understand gpt-3.
\newblock \emph{Proceedings of the National Academy of Sciences},
  120(6):e2218523120.

\bibitem[{Cai et~al.(2023)Cai, Haslett, Duan, Wang, and
  Pickering}]{cai2023does}
Zhenguang~Garry Cai, David~A Haslett, Xufeng Duan, Shuqi Wang, and Martin~John
  Pickering. 2023.
\newblock Does chatgpt resemble humans in language use?

\bibitem[{Chang and Bergen(2022)}]{chang2022word}
Tyler~A Chang and Benjamin~K Bergen. 2022.
\newblock Word acquisition in neural language models.
\newblock \emph{Transactions of the Association for Computational Linguistics},
  10:1--16.

\bibitem[{Hu et~al.(2022)Hu, Floyd, Jouravlev, Fedorenko, and
  Gibson}]{hu2022fine}
Jennifer Hu, Sammy Floyd, Olessia Jouravlev, Evelina Fedorenko, and Edward
  Gibson. 2022.
\newblock A fine-grained comparison of pragmatic language understanding in
  humans and language models.
\newblock \emph{arXiv preprint arXiv:2212.06801}.

\bibitem[{Lampinen et~al.(2024)Lampinen, Dasgupta, Chan, Sheahan, Creswell,
  Kumaran, McClelland, and Hill}]{lampinen2024language}
Andrew~K Lampinen, Ishita Dasgupta, Stephanie~CY Chan, Hannah~R Sheahan,
  Antonia Creswell, Dharshan Kumaran, James~L McClelland, and Felix Hill. 2024.
\newblock Language models, like humans, show content effects on reasoning
  tasks.
\newblock \emph{PNAS nexus}, 3(7):pgae233.

\bibitem[{Mayo et~al.(2014)Mayo, Schul, and Rosenthal}]{mayo2014if}
Ruth Mayo, Yaacov Schul, and Meytal Rosenthal. 2014.
\newblock If you negate, you may forget: negated repetitions impair memory
  compared with affirmative repetitions.
\newblock \emph{Journal of Experimental Psychology: General}, 143(4):1541.

\bibitem[{Nguyen(2024)}]{nguyen2024human}
Jeremy~K Nguyen. 2024.
\newblock Human bias in ai models? anchoring effects and mitigation strategies
  in large language models.
\newblock \emph{Journal of Behavioral and Experimental Finance}, 43:100971.

\bibitem[{Schubert et~al.(2024)Schubert, Jagadish, Binz, and
  Schulz}]{schubert2024context}
JA~Schubert, AK~Jagadish, M~Binz, and E~Schulz. 2024.
\newblock In-context learning agents are asymmetric belief updaters (arxiv:
  2402.03969). arxiv.

\bibitem[{Suri et~al.(2024)Suri, Slater, Ziaee, and Nguyen}]{suri2024large}
Gaurav Suri, Lily~R Slater, Ali Ziaee, and Morgan Nguyen. 2024.
\newblock Do large language models show decision heuristics similar to humans?
  a case study using gpt-3.5.
\newblock \emph{Journal of Experimental Psychology: General}.

\bibitem[{Warstadt and Bowman(2022)}]{warstadt2022artificial}
Alex Warstadt and Samuel~R Bowman. 2022.
\newblock What artificial neural networks can tell us about human language
  acquisition.
\newblock In \emph{Algebraic structures in natural language}, pages 17--60. CRC
  Press.

\bibitem[{Zang et~al.(2023)Zang, Beltr{\'a}n, Wang, Rol{\'a}n, and
  de~Vega}]{zang2023negation}
Anqi Zang, David Beltr{\'a}n, Huili Wang, Katia Rol{\'a}n, and Manuel de~Vega.
  2023.
\newblock The negation-induced forgetting effect remains even after reducing
  associative interference.
\newblock \emph{Cognition}, 235:105412.

\end{thebibliography}

\end{document}